# IMPROVED DIMENSIONALITY REDUCTION OF VARIOUS DATASETS USING NOVEL MULTIPLICATIVE FACTORING PRINCIPAL COMPONENT ANALYSIS (MPCA)

Chisom Ogbuanya

Department of Electronic Engineering

University of Nigeria, Nsukka

Nigeria

**Abstract**

Principal Component Analysis (PCA) is known to be the most widely applied dimensionality reduction approach. A lot of improvements have been done on the traditional PCA, in order to obtain optimal results in the dimensionality reduction of various datasets. In this paper, we present an improvement to the traditional PCA approach called Multiplicative factoring Principal Component Analysis (MPCA). The advantage of MPCA over the traditional PCA is that a penalty is imposed on the occurrence space through a multiplier to make negligible the effect of outliers in seeking out projections. Here we apply two multiplier approaches, total distance and cosine similarity metrics. These two approaches can learn the relationship that exists between each of the data points and the principal projections in the feature space. As a result of this, improved low-rank projections are gotten through multiplying the data iteratively to make negligible the effect of corrupt data in the training set. Experiments were carried out on YaleB, MNIST, AR, and Isolet datasets and the results were compared to results gotten from some popular dimensionality reduction methods such as traditional PCA, RPCA-OM, and also some recently published methods such as IFPCA-1 and IFPCA-2.

## 1.0   Introduction

Principal component analysis (PCA) has always been known as a reliable method of reducing the dimension of data collected [1], and other things such as pattern recognition and data compression [2]. It is crucial, in most cases, to reduce the dimensionality of any kind of data collected before processing to give lower computational cost and memory requirements, without any loss in the quality of the data being processed [1]. Whether the data consists of images for hyper spectral imaging, or industrial data for industrial purposes, dimensionality reduction of such data is crucial if not compulsory because their size/dimensionality can range from thousands to hundreds of thousands [3]. Traditional PCA has been globally recognized as an easy and accurate dimensionality reduction technique [4]. It develops the best approximation in the subspace, in terms of observations, in a way that is least-square. It achieves this by calculating singular value decomposition of the first dataset [4]. As a result of its quadratic error criterion, PCA is sensitive to outliers, in the face of datasets.

The performance of PCA is not at its best when a minimum number of outliers are involved. This is the reason a lot of research has gone into the improvement or modification of PCA for optimal performance of dimensionality reduction of high dimension data or any amount/kind of data collected.

2.0  **Related Works**

Improved PCA methods such as Optimal mean robust principal component analysis (RPCA-OM) [5] subtracts the average in a given dataset by integrating the average, automatically into the objective function of the dimensionality reduction. Ganaa et al in [6] applied instance factoring PCA (IFPCA) which involved applying a scaling-factor as a penalty that is forced on the instance space, in order, to hold down the effect of outliers in going after projections. Improved PCA (IPCA) is almost the same with PCA except that it applies the Shannon information theory for developing/upgrading the PCA algorithm. In [7] graph-Laplacian PCA (gLPCA) uses graph structures to learn a low dimensional representation of data. In [8] PCA was improved using a bootstrap resampling method, in order, to solve the problem of small sample during the process of data mean centering. [1] Developed a High-dimensional robust PCA to solve the problem of quadratic error criterion whenever outliers occur as PCA is being used for dimensionality reduction. Deterministic high-dimensional robust PCA (DHR-PCA) is proposed in [9] to also deal with the issue of outliers by reducing the weight of all of the observations such that the weight of outliers automatically reduces quicker than the other observations, so that the outliers end up having a diminishing effect on the sample co-variance matrix. In [10], block PCA was improved by when the transform results were ensured to have the maximum variance where necessary. In [11], the characteristics gotten from Deep Belief Networks were used to put an end to the data processing challenges in industrial control system. Their work was strategically focused on solving the problems of missing values in dimensionality reduction. A strategy for power quality data compression was proposed in [12], which has its basis on PCA in wavelet domain and layered coding. Sparse PCA was improved in [13], by the sparse PCA being made adaptive, such that the same sparsity pattern can be gotten across all principal components. [14] Solved the problem of sensitivity to outliers by proposing a new method that introduces the self-paced learning mechanism into probabilistic principal component analysis. Here, the self-paced probabilistic PCA deploys an iterative procedure to search out the optimal projection vectors, and then dispose of the outliers.

In this work, we propose a method that attempts to accommodate the sensitivity of PCA to noise. The proposed method is called Multiplicative factoring PCA. In this MPCA, a multiplier that suppresses the impact of outliers or noise in pursuing projections is applied. Two major paths are followed in this work: angular metrics and total distance are applied spatially to iteratively learn the correlation between one of all the instances and the principal projection in the feature space. As a result of this, the two major paths can differentiate between genuine data and noise. The main contributions of this paper are summarized below:

1. We propose an improvement on the traditional PCA model by introducing a multiplier, in order, to suppress the impact of outliers.

2. We also propose two major paths: angular metrics and total distance. These metrics iteratively check the essentiality of each instance by understanding and comparing the relationship between each instance and the principal projection in the feature space.

Experiments on YaleB, MNIST, AR, and Isolet datasets prove that our method performs better than traditional PCA, RPCA, and IFPCA.

The rest of this paper is organized as follows: section 2 explains the proposed method thoroughly. Section 3 presents the experiments, results and complexity analysis, while conclusions and recommendations are given in section 4.

## 3.0 The Proposed Method

To explicitly explain our proposed method, we begin by taking note of the objective function of PCA [5]:

$$min_{w^Tw=1} \sum_{i=1}^{n}(x_i - ww^Tx_i) = min\|X - ww^TX\|_2^2 \qquad (1)$$

Where $\{w\}_{j=1}^{d}$ is a subset of orthogonal projection vectors in $\mathbb{R}^m$ and the set of data points $\{x_i\}_{i=1}^{h}$ is zero-mean $m$-dimensional data points. It can be observed that, PCA applies a least square approach to reduce the sum-distance between the original dataset $X$ and the reconstructed dataset $ww^TX$. This will geometrically force the projection vector $w$ to move through the data points that are most dense to reduce the sum-distance. (This is as shown in figure 1). In summary, $u_1$ is the first principal projection vector $u_1$. From this understanding, we verify the essence of examples by putting into consideration the relationship between each example and the principal projection. That is, the nearer an example to the projection vector $u_1$, the more essential the example in chasing after projection.

Therefore, we stretch formula (1) to also consist of a *multiplicative factor*. This factor imposes a penalty on the example space to suppress the influence of noise in datasets that are not complete. The following is our proposed function:

$$\min_{D_V}\|XD - VV^TXD\|_F^2 = \max_{D_V} V^TDX^TXD_V \qquad (2)$$

$$s.t. V^T.V = 1$$

Where $V$ is a vector of sample space and $D = diag(d_1, d_2, \ldots d_n)$ is a diagonal matrix that tests the essence of each example in $X$. With this penalty, we are therefore, chasing after a projection $Z = D$ with $Z^TZ = 1$ that puts into consideration the effect of examples. For instance, if a lower multiplicative – factor $d_i$ is assigned to the projection $Z$, the component of sample space $Z_i$ is suppressed, which implies that the corresponding sample $x_i$ makes only a little contribution to the projection $Z$.

To ensure that the constraint in formula (2) is maintained, we introduce the Lagrange multiplier ($\lambda$) and obtain partial derivatives w.r.t V, in order to have:

$$X^T X D D V = \lambda V \quad (3)$$

It can be observed that formula (3) is a standard eigenvalue problem.

Applying mathematics, there is a direct relationship between PCA and Singular Value Decomposition (SVD) [5] when the components of PCA are calculated from the covariance matrix. The formula for SVD of $X$ is as given below:

$$X = u \sum V^T \quad s.t. \ u^T u = I_r, V^T V = I_r \quad (4)$$

For our proposed method, V=DV, where V is the set of $r$ projections of V. Therefore, the projection $u$ in feature space can be gotten as follows:

$$u = XDV\varepsilon^{-1} \quad (5)$$

Where the low dimension feature space $u$ is gotten with an infusion of sample factors, which vary from the classical PCA. In this form, MPCA can study a low dimensional subspace from both sample and feature spaces of a dataset for better performance.

### 3.1 Methods of Building Matrix D

Here we describe the relationship between the multiplicative-factor $D$ and the principal projection $u_1$ using two approaches: total distance and cosine similarity metrics. Both of them can be gotten geometrically.

**Total Distance Metric**

The method first applies total distance metric, in order to, learn iteratively the relationship that exists between each occurrence and the principal projection $u_1$. The total distance of an occurrence is stated as the square sum of the distances between the coordinate of each occurrence and the coordinates of each of the other occurrence in the training set to the projection $u_1$. The total distance of an occurrence is a normal method to verify its essence within the set. Geometrically, the total distance of occurrence $x_i$ which is within the cluster just as the occurrence will be larger than occurrence $x_j$ because of the effect of the multiplicative factor. Therefore, occurrence $x_i$ is in a better position to be an outlier or corrupt occurrence than $x_j$. Geometrically, therefore, the coordinate of occurrence squared to the projection $u_1$ is computed through:

$$s_i = u_1^T x_i^2 \quad (6)$$

The next thing is to obtain the formula for $d_i$ through total distance metric as follows:

$$d_i = \sum_{i,j=1}^{n} (s_i - s_j)^2 \qquad (7)$$

Therefore, the larger the $d_i$, the more $x_i$ is in a position to be a corrupt occurrence or noisy, and so its essence will be multiplied, in order, to make negligible its effect on the projection.

**Angular metrics**

The second thing the method does is that it applies cosine similarity metric to construct the multiplicative-factor D. This also iteratively ensures to learn the angle relationship set and the principal projection $u_1$. Therefore, by making formula (6) normal, the angle between each occurrence and the principal projection $u_1$ is stated definitively as follows:

$$Cos\ \theta_i = \frac{u_1^T x_i^2}{\|u_1\| * \|x_i\|} \qquad (8)$$

As can be seen from formula (8), a larger $Cos\ \theta_i$ means a lesser angle $\theta_i$ between instance $x_i$ and the principal projection $u_1$ and vice versa. Also, angle ϕ of occurrence $x_j$ is almost lesser than angle θ of instance $x_i$. Therefore, $x_j$ will be put into consideration as more essential in seeking out best projections than $x_i$ which might be corrupt occurrence. Noting that $d_i$ is a negative factor, we evaluate $d_i$ through the similarity metric as follows:

$$d_i = \frac{1}{abs(Cos\ \theta_i) + \varepsilon} \qquad (9)$$

Where $\varepsilon = 0.0001$ is a parameter to prevent $d_i$ from getting close to infinity. Multiplying the data iteratively, by applying the two approaches listed above, the impact of noise in the training set will be minimized appropriately, so as to achieve improved low-rank projections. Below in Figure 1 is the algorithmic description of the proposed MPCA method:

|    | **Algorithm of the proposed MPCA method** |
|----|-------------------------------------------|
| 1  | Input: Training set X |
| 2  | Output: The projection vector V |
| 3  | Parameters: $\varepsilon$ |
| 4  | Initialize: Initialize D as an identity matrix |
| 5  | While not converged do |
| 6  | Obtain V based on formula (3) |
| 7  | Obtain U based on formula (5) |
| 8  | Update D based on formulas (7) or (9) |
| 9  | Compute loss from formula (2) |
| 10 | End while |

**Figure 1:** Algorithm of the proposed MPCA method

## 4.0 Experiments and Discussion

We carried out experiments on YaleB, MNIST, AR, and Isolet datasets using our proposed MPCA algorithm and with also some other dimensionality reduction methods such as PCA, RPCA-OM, IFPCA-1, and IFPCA-2. These were done, in order to, show the efficiency of our proposed MPCA algorithm.

## 4.1: Parameter Settings

For each of the datasets, we randomly sampled 60%, and 80% of the occurrences for training and testing respectively. The parameters of PCA [15], RPCA-OM [16], IFPCA-1 [6], and IFPCA-2 [6] were set as stated in their literatures. For the MPCA, we set the KNN parameter **k** to 5, and also for the other comparative methods, so as to ensure fair comparison. The use of k-nearest neighbor (KNN) classifier was made for the classifications. We denote results for our design as MPCA-1 and MPCA-2, where MPCA-1 and MPCA-2 are names for cosine similarity and total distance metrics respectively. We ran each experiment 10 times and we took note of the average classification accuracies, which represent optimal dimensions and standard deviations for the different approaches (MPCA-1 and MPCA-2).

## 4.2: Discussion of Results and Analysis

In this section, we hereby discuss the results gotten for each approach on the 4 datasets used for our experiments, and also carry out analysis of each of the results.

**Face Recognition:** We carried out experiments on 2 face datasets, YaleB, and AR dataset, to validate the effectiveness of our proposed algorithm on face recognition. The YaleB dataset consists of 5760 single-light-source images of 10 individuals, each of the individuals under 576 viewing conditions (9 poses * 64 illumination conditions) [17]. The results obtained when the different methods were used on this dataset are shown in Table 1 with optimal results in bold. We can observe from Table 1 that MPCA-1 and MPCA-2 both have better accuracies than all the comparative methods. For optimal dimensions, MPCA-1 and MPCA-2 begot the highest optimal dimensions in both 60% and 80% samples when compared to the other comparative methods. From Table 1, we can see that for the 60% sample, MPCA-1 and MPCA-2 outperforms PCA, RPCA-OM, IFPCA-1, and IFPCA-2 by approximately 4.7%, 2.32%, 1.58%, and 0.83% respectively. Also, MPCA-1 and MPCA-2 got the lowest variances in both samples, which demonstrates that their better performance is consistent.

**Table 1:** Mean classification accuracies ± standard deviation (%) and optimal dimensions (in parentheses) of the various methods on the Yale B dataset

| Dataset | YALEB | | | | | |
|---|---|---|---|---|---|---|
| Method | MPCA-1 | MPCA-2 | PCA | RPCA-OM | IFPCA-1 | IFPCA-2 |
| **60%** | 71.19 ±0.92 (24) | 70.95 ± 0.94 (792) | 66.49 ± 1.08 (140) | 68.87 ± 1.39 (458) | 69.61 ± 0.92 (481) | 70.36 ± 0.56 (210) |
| **80%** | 80.00 ± 0.00 (26) | 78.12 ± 0.39 (800) | 69.9 ± 1.39 (137) | 74.18 ± 1.28 (508) | 75.78 ± 0.95 (500) | 78.04 ± 0.74 (351) |

**YALE B (80% TRAINING)**

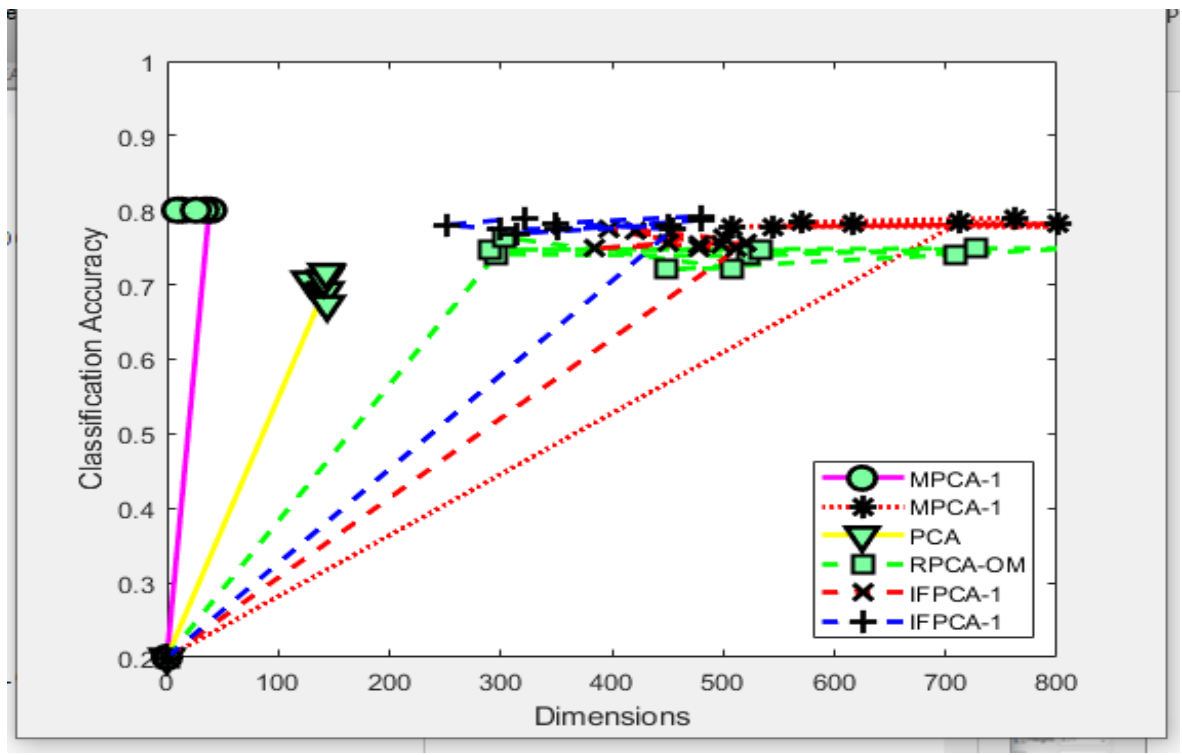

**Figure 2:** Classification accuracies against the variations of optimal dimensions in the Yale B dataset for the different methods.

The AR face dataset is made up of 126 people (over 4,000 color images) [18]. Images feature frontal view faces with different facial expressions, illumination conditions and occlusions. The results for the different methods applied in carrying out experiments on the AR dataset are shown in Table 2 with optimal results in bold. From Table 2, we can also observe that MPCA-1 and MPCA-2 have the lowest variance in both samples, further validating the stability of the performance of our proposed algorithm.

These results obtained from the face recognition experiment have shown that the proposed methods have the highest performances than all the other comparative methods. The results obtained prove the same. Figure 3 demonstrates the classification accuracies of each approach against the variation of dimensions. It is clear from Figure 2 that MPCA-1, MPCA-2, RPCA-OM, PCA, IFPCA-1 and IFPCA-2 reach consistent performances in higher dimensions above 10.

**Table 2:** Mean classification accuracies ± standard deviation (%) and optimal dimensions (in parentheses) of the various methods on the AR dataset

| Dataset | AR | | | | | |
|---|---|---|---|---|---|---|
| Method | MPCA-1 | MPCA-2 | PCA | RPCA-OM | IFPCA-1 | IFPCA-2 |
| **60%** | 70.92±1.01 (133) | 71.03 ± 0.97 (141) | 69.40 ± 2.19 (255) | 70.02 ± 1.05 (265) | 70.52 ± 1.10 (261) | 70.34 ± 0.91 (287) |
| **80%** | 80.54±0.48 (133) | 80.87 ± 0.96 (151) | 78.60 ± 1.11 (259) | 79.53 ± 1.37 (278) | 80.44 ± 1.58 (264) | 79.61 ± 0.77 (288) |

**AR (80% TRAINING)**

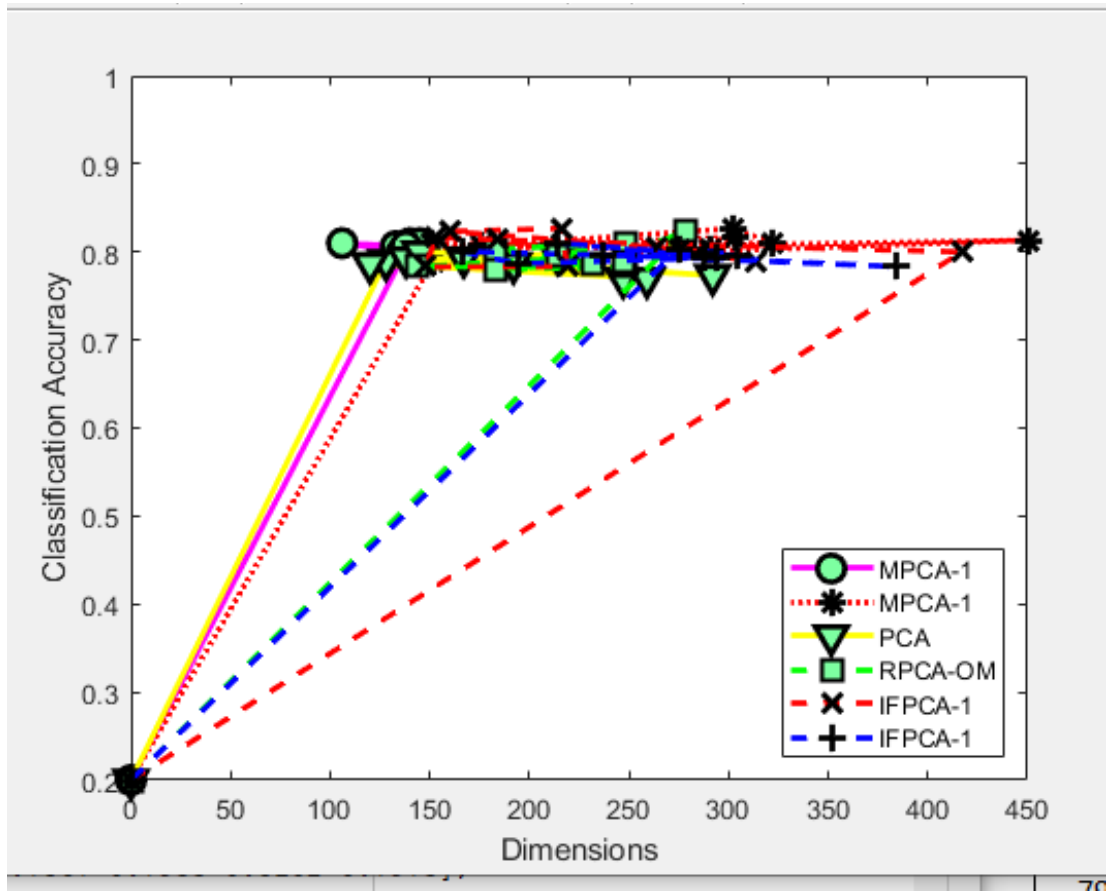

**Figure 3:** Classification accuracies against the variations of optimal dimensions in the AR dataset for the different methods.

**Handwritten Digit Recognition:** We also carried out experiments on the MNIST dataset, in order to, further validate the efficient performance of our design. The MNIST dataset is made up of handwritten digits, which is a training set of 60,000 examples, and a testing set of 10,000 examples [19].

MPCA-1 and MPCA-2 also performed better than all the other comparative methods in written digit recognition for both training samples of the MNIST dataset, as can be seen in Table 3. From Table 3, for the 60% sample, MPCA-1 and MPCA-2 has an approximate digit recognition accuracy of 0.87%, 0.83%, 0.54%, and 0.58% more than PCA, RPCA-OM, IFPCA-1, and IFPCA-2 respectively. MPCA-1 went ahead to get the best optimal dimensions of 26 and 38 in the 60% and 80% samples respectively.

**Table 3:** Mean classification accuracies ± standard deviation (%) and optimal dimensions (in parentheses) of the various methods on the MNIST dataset

| Dataset | MNIST | | | | | |
|---|---|---|---|---|---|---|
| Method | MPCA-1 | MPCA-2 | PCA | RPCA-OM | IFPCA-1 | IFPCA-2 |
| **60%** | 92.85±0.19 (26) | 92.42 ± 0.28 (33) | 91.98 ± 0.46 (25) | 92.02 ± 0.38 (33) | 92.31 ± 0.31 (34) | 92.27 ± 0.39 (40) |
| **80%** | 93.93±0.41 (38) | 93.61 ± 0.33 (38) | 93.16 ± 0.87 (35) | 93.23 ± 0.84 (35) | 93.77 ± 0.84 (49) | 93.31 ± 0.97 (51) |

**MNIST (80% TRAINING)**

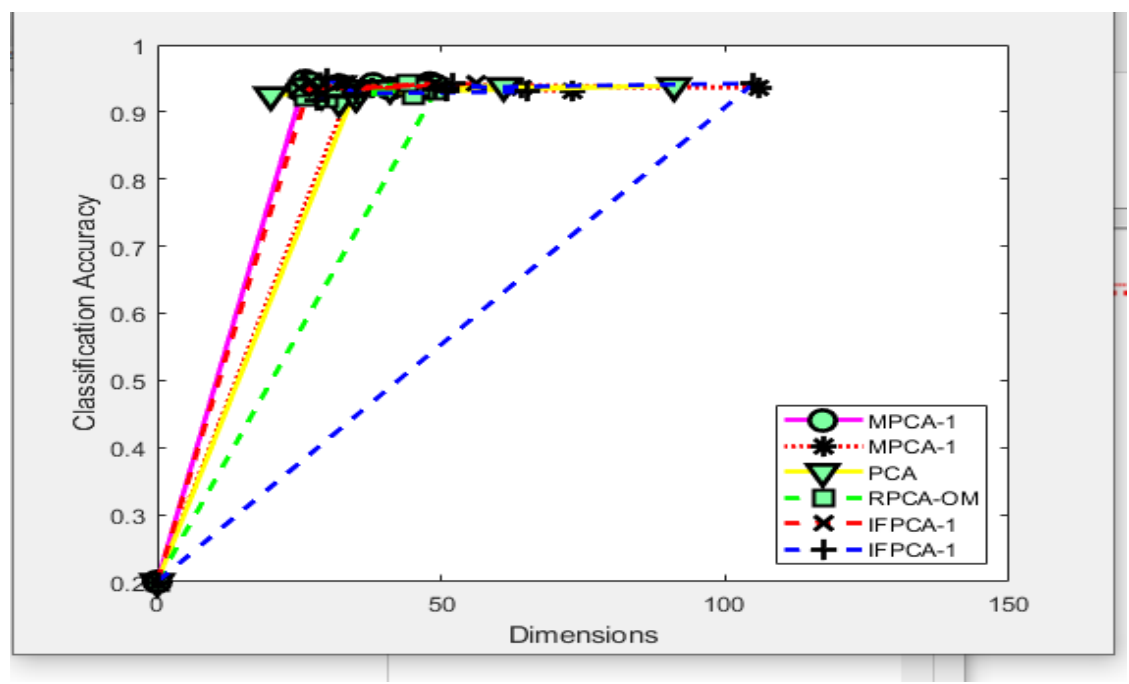

**Figure 4:** Classification accuracies against the variations of optimal dimensions in the MNIST dataset for the different methods.

**Voice Recognition:** Experiments on the Isolet dataset were also carried out to finally prove the effectiveness of our algorithm. The Isolet dataset was generated in this way: 150 people pronounced the name of each letter of the alphabet two times. Therefore, there are 52 training examples from each of the speakers [20].

From Table 4 we can observe that MPCA-1 and MPCA-2 outperformed all the comparative methods in voice recognition for both training samples of the Isolet dataset. For the 60% sample, MPCA-1 has a voice recognition accuracy of 2.03%, 1.25%, 0.5%, and 1.48% more than PCA, RPCA-OM, IFPCA-1, and IFPCA-2 respectively. MPCA-1 went ahead to obtain the best optimal dimensions of 33 and 41 in the 60% and 80% sample respectively.

**Table 4:** Mean classification accuracies ± standard deviation (%) and optimal dimensions (in parentheses) of the various methods on the ISOLET dataset

| Dataset | ISOLET | | | | | |
|---|---|---|---|---|---|---|
| Method | MPCA-1 | MPCA-2 | PCA | RPCA-OM | IFPCA-1 | IFPCA-2 |
| **60%** | 92.12±0.90 (33) | 91.03 ± 0.71 (30) | 90.09± 0.41 (22) | 90.87 ± 0.57 (39) | 91.62 ± 1.15 (81) | 90.64 ± 0.82 (30) |
| **80%** | 93.85±0.77 (41) | 93.94 ± 0.89 (40) | 92.63 ± 0.00 (23) | 92.85 ± 0.76 (47) | 92.89 ± 0.85 (96) | 93.24 ± 0.83 (32) |

**ISOLET (80% TRAINING)**

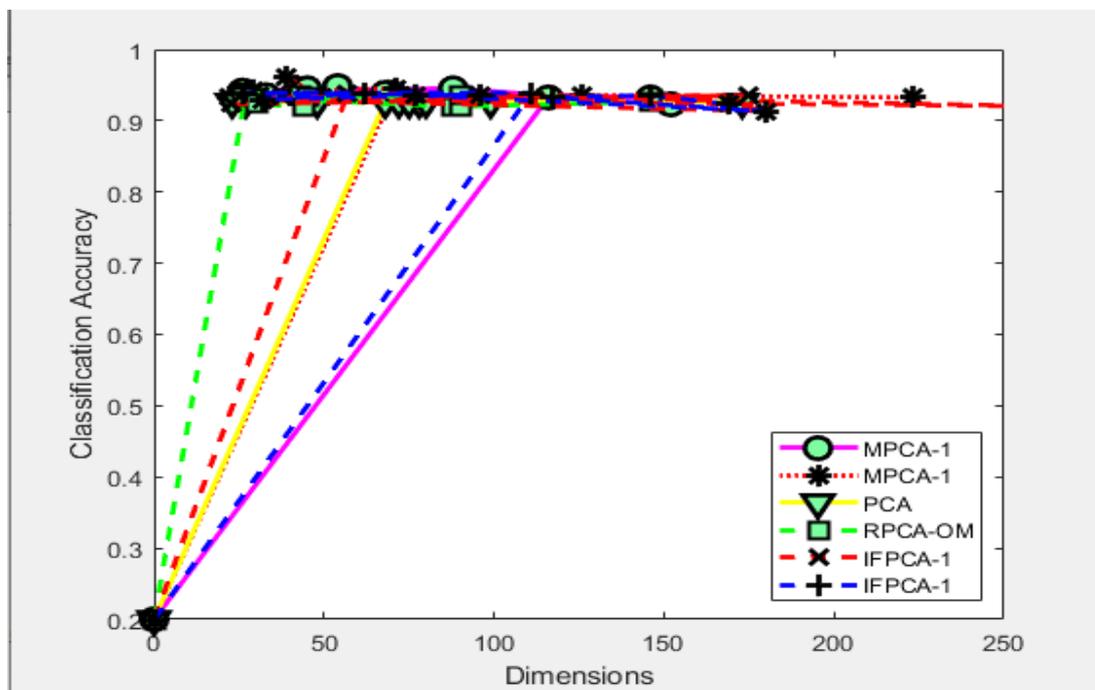

**Figure 5:** Classification accuracies against the variations of optimal dimensions in the ISOLET dataset for the different methods.

From Figure 5, we can observe the classification accuracies of each of the methods against the differences in dimensions. We can observe that the methods demonstrate consistent performances in dimensions above 20, however, MPCA-1 and MPCA-2 takes the lead. In essence, the stability in the performances of MPCA-1 and MPCA-2 showed their effectiveness in seeking out, in an improved way, the internal details of each of the datasets. Our proposed methods can also be commended for their abilities to obtain the lowest variances than the comparative methods.

**Complexity Analysis:** Here we make comparisons of the computational times of the proposed methods to the other comparative methods. All the algorithms were implemented in MATLAB R2016b Version 9.1.0.441655 64-bit using a desktop computer with Intel ® Core ™ i5-7500 CPU @3.40GHz with 8.00GB memory and Windows 7 operating system environment. The proposed algorithm has a convergence that relies on the essence evaluation diagonal matrix *D*. The formular for calculating the amount of time of an eigenvalue problem on a training set X of size $m*n$ is $O(m^3)$. This implies that a complexity of $O(m^3)$ is needed by the proposed algorithm to calculate the projection vector *p*, considering that our algorithm is an eigenvalue problem. As for the internal loop, if it uses **k** number of iterations in seeking out D for convergence to be reached, the complexity is $O(kmn)$. Therefore, the overall complexity of the algorithm becomes $O(t(m^3+kmn))$, where *t* is the number of iterations of the external loop. In Table 5, we can see the computation time for each method on all three datasets.

**Table 5:** Approximate Computation time in seconds for each method on all three datasets.

| Dataset | YaleB | | AR | | MNIST | | Isolet | |
|---|---|---|---|---|---|---|---|---|
| | **Training** | **Testing** | **Training** | **Testing** | **Training** | **Testing** | **Training** | **Testing** |
| **MPCA-1** | $1.92 \times 10^{-1}$ | $1.81 \times 10^{-2}$ | $1.81 \times 10^{-1}$ | $1.69 \times 10^{-1}$ | 1.62 | $1.22 \times 10^{-1}$ | $2.75 \times 10^{-2}$ | $1.64 \times 10^{-2}$ |
| **MPCA-2** | $1.70 \times 10^{-1}$ | $1.58 \times 10^{-2}$ | $1.51 \times 10^{-1}$ | $1.47 \times 10^{-2}$ | 3.92 | $1.15 \times 10^{-1}$ | $2.72 \times 10^{-2}$ | $1.61 \times 10^{-3}$ |
| **PCA** | $7.55 \times 10^{-2}$ | $6.43 \times 10^{-3}$ | $6.78 \times 10^{-2}$ | $5.67 \times 10^{-3}$ | $3.31 \times 10^{-1}$ | $2.20 \times 10^{-2}$ | $2.88 \times 10^{-2}$ | $1.76 \times 10^{-3}$ |
| **RPCA-OM** | $1.22 \times 10^{-1}$ | $1.36 \times 10^{-2}$ | $1.10 \times 10^{-1}$ | $1.49 \times 10^{-2}$ | 1.67 | $1.43 \times 10^{-1}$ | $1.12 \times 10^{-1}$ | $1.76 \times 10^{-2}$ |
| **IFPCA-1** | $1.81 \times 10^{-1}$ | $1.74 \times 10^{-2}$ | $1.70 \times 10^{-1}$ | $1.59 \times 10^{-2}$ | 1.58 | $1.47 \times 10^{-1}$ | $2.10 \times 10^{-2}$ | $2.66 \times 10^{-3}$ |
| **IFPCA-2** | $1.57 \times 10^{-1}$ | $1.46 \times 10^{-2}$ | $1.46 \times 10^{-1}$ | $1.35 \times 10^{-2}$ | 3.71 | $2.60 \times 10^{-1}$ | $2.10 \times 10^{-2}$ | $2.57 \times 10^{-3}$ |

## 5.0 Conclusion and Recommendation

We presented in this paper a novel dimensionality reduction method called multiplicative factoring PCA. This is an improvement to PCA which involves a multiplier that imposes a penalty on the occurrence space, so as to make negligible the effect of corrupt data in seeking out projections. Two approaches, cosine similarity and total distance metrics are applied geometrically to learn iteratively the relationship that exists between each occurrence and the principal projection. Extensive experiments were carried out on popular datasets such as YaleB, MNIST, AR, and Isolet to demonstrate the extent of the improvement of MPCA, in both dimensionality reduction and classification tasks. This improvement is made more obvious by the comparison made between MPCA and the state-of-the art methods such as PCA, RPCA-OM, IFPCA-1 and IFPCA-2. Our method proved to be more unbiased with corrupt data than other comparative methods. For future work, this method will be applied to enhanced graph embedding.